\documentclass[letterpaper, 10 pt, conference]{ieeeconf}  % Comment this line out if you need a4paper

\IEEEoverridecommandlockouts                              % This command is only needed if 
\usepackage{cite}
\usepackage{amsmath,amssymb,amsfonts}
\usepackage{graphicx}
\usepackage{subcaption}
\usepackage{textcomp}
\usepackage{color}
\usepackage{hyperref}
\usepackage[noabbrev]{cleveref}
\usepackage{multirow}
\usepackage{tabularray}
\usepackage[T1]{fontenc}
\usepackage{algorithm} 
\usepackage[noend]{algpseudocode}
\usepackage{fancyhdr}

\title{\LARGE \bf
Bevel-Tip Needle Deflection Modeling, Simulation, and Validation in Multi-Layer Tissues
}

\author{Yanzhou Wang$^{*1}$, Lidia Al-Zogbi$^{*1}$, Guanyun Liu$^{2}$, Jiawei Liu$^{3}$, Junichi Tokuda$^{4}$,\\ Axel Krieger$^{1}$, and Iulian Iordachita$^{1}$  % <-this % stops a space
\thanks{This work is supported by NIH R01R01EB020667, 1R01EB025179, 1R01CA235134, and in part by a collaborative research agreement with the Multi-Scale Medical Robotics Center in Hong Kong}% <-this % stops a space
\thanks{$^{1}$Yanzhou Wang, Lidia Al-Zogbi, Axel Krieger, and Iulian Iordachita are with the Department of Mechanical Engineering and the Laboratory of Computational Sensing and Robotics, Johns Hopkins University, Baltimore, MD, USA
  {\tt\small ywang521@jh.edu}}%
\thanks{$^{2}$Guanyun Liu is with the Department of Mechanical and Aerospace Engineering, University of Florida, Ganesville, USA}%
\thanks{$^{3}$Jiawei Liu is with the Laboratory of Computational Sensing and Robotics, Johns Hopkins University, Baltimore, MD, USA}%     
\thanks{$^{4}$Junichi Tokuda is with the Department of Radiology, Brigham and Women’s Hospital and Harvard Medical School, Boston, MA, USA}
}

\begin{document}
\onecolumn
\noindent This work has been submitted to the IEEE for possible publication. Copyright may be transferred without notice, after which this version may no longer be accessible.

\twocolumn

\maketitle

\begin{abstract}
Percutaneous needle insertions are commonly performed for diagnostic and therapeutic purposes as an effective alternative to more invasive surgical procedures. However, the outcome of needle-based approaches relies heavily on the accuracy of needle placement, which remains a challenge even with robot assistance and medical imaging guidance due to needle deflection caused by contact with soft tissues. In this paper, we present a novel mechanics-based 2D bevel-tip needle model that can account for the effect of nonlinear strain-dependent behavior of biological soft tissues under compression. Real-time finite element simulation allows multiple control inputs along the length of the needle with full three-degree-of-freedom (DOF) planar needle motions. Cross-validation studies using custom-designed multi-layer tissue phantoms as well as heterogeneous chicken breast tissues result in less than 1mm in-plane errors for insertions reaching depths of up to 61 mm, demonstrating the validity and generalizability of the proposed method.
% The model is successfully validated using a flexible bevel-tipped needle inserted into custom-designed multi-layer tissue phantoms as well as heterogeneous chicken breast tissues, achieving in-plane needle errors smaller than 1 mm. The model is also cross-validated achieving errors also smaller than 1 mm, successfully demonstrating its generalization capabilities. 
\end{abstract}

\section{Introduction}
\label{sec:introduction}

\begin{figure*}[t]
  \centering
  \includegraphics[width=\textwidth]{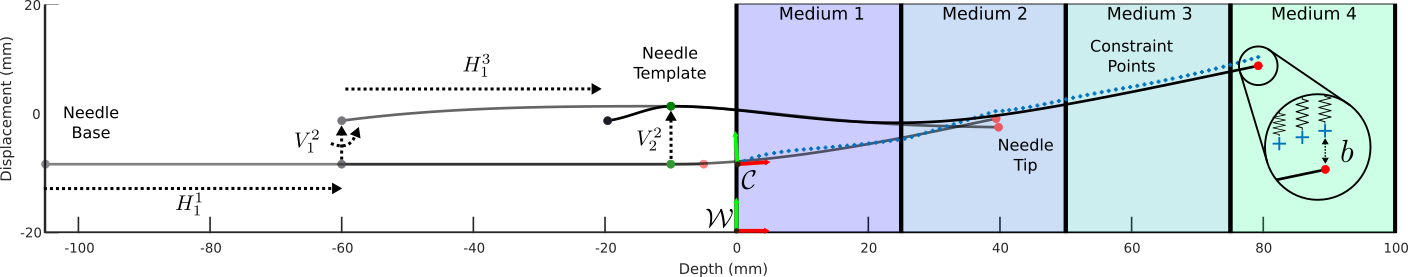}
  \caption{Bevel tip needle insertion simulation with four layers of Ogden-type soft medium. Needle base (black dot) controls horizontal ($H_1^m$) and vertical displacements and slope ($V_1^m$). A horizontally-fixed needle template (green dot) controls vertical displacement and prescribes zero slope ($V_2^m$). Constraint points (blue cross) are placed dynamically, and simulate a bevel effect at the needle tip (red dot). Input steps $m\in \left\{ 1, 2, 3 \right\}$.}
  \label{fig:model_schematic}
\end{figure*}

Percutaneous needle interventions play a fundamental role in both the diagnosis and treatment of many diseases, in particular prostate cancer and breast cancer. According to the American Cancer Society's update on cancer statistics, prostate cancer incidence has increased by 3\% annually from 2014 through 2019, and breast cancer incidence rates have risen in most of the past four decades~\cite{Siegel2020, Giaquinto2022, Siegel2023}. To improve surgical outcomes, bevel-tip needle insertions have been heavily researched over the past two decades due to their extensive use in minimally invasive percutaneous procedures such as biopsies~\cite{Taschereau2000, Robertson2011, Wu2011}, brachytherapy~\cite{Damore2000, Wan2005, Podder2006}, as well as spinal injections~\cite{Abolhassani2007, Gao2012, Van2012, Van2014}.

Bevel-tip needle insertions are modeled using kinematic descriptions originating from the field of vehicle dynamics, where needle motions are formulated from a series of nonholonomic constraints~\cite{Park2005, Webster2006, Alterovitz2008}. Although kinematic models can simplify needle path planning and shape prediction, their application is limited to cases where the needle is highly flexible relative to the surrounding soft tissues, and the interaction between needle shaft and soft tissues is negligible. Additionally, needle steering is typically restricted to insertion and axial rotation, while avoiding any lateral needle motion that can cause the flexible needle to bend due to soft tissue interactions, further limiting the overall generalizability of such approach.

One way to account for the interaction between the needle's shaft and soft tissue is through force modeling~\cite{DiMaio2005_1, DiMaio2005_2, Abolhassani2006, Glozman2007, Abayazid2013, Roesthuis2011, Adagolodjo2019, Wang2023_1}. In planar cases, the lateral interaction between a flexible needle and soft tissue is typically represented by a linear elastic beam for the needle shaft, and distributed linear springs along the needle shaft for the tissue~\cite{DiMaio2005_1, DiMaio2005_2, Abolhassani2006, Glozman2007, Abayazid2013}. For example, DiMaio and Salcudean~\cite{DiMaio2005_1, DiMaio2005_2} present a mechanics-based model based on linear needle bending model coupled with a linear tissue material, and they use needle base control to guide a symmetric-tip needle to a desired location. Glozman and Shoham~\cite{Glozman2007} use a distributed linear spring model to represent tissue reaction force, and their method relies on X-ray feedback to close the control loop. Roesthuis~\textit{et al.} extend the formulation to bevel tip needle by considering a transverse tip force generated by soft tissues during needle insertion~\cite{Roesthuis2011}, and Abayazid~\textit{et al.} further enrich the formulation by considering double-bend shape of a bevel-tip needle~\cite{Abayazid2013}. 

Notably, work done on symmetric-tip needles is primarily focused on lateral base motion to achieve needle steering; on the other hand, bevel-tip needle steering is typically achieved only by changing the bevel orientation via axial rotation, while largely ignoring lateral motion of the needle. However, in reality, both lateral base motion and needle tip bevel can be used for steering, albeit with different working principles. Furthermore, a common feature of previous work is the use of largely simplified linear tissue behavior model. Biological soft tissues exhibit highly nonlinear stress-stretch behavior, which cannot be captured by such models~\cite{Humphrey2004, Singh2021}. The lack of a realistic tissue model also makes the choice of tissue linearity measure rather arbitrary. Adagolodjo~\textit{et al.} address tissue nonlinearity in symmetric needle insertions using the SOFA framework~\cite{Adagolodjo2019}, an open-source interactive finite element physics simulator designed to provide real-time computational capabilities~\cite{SOFA_2012, SOFA_2017}. However, no model validation is provided in~\cite{Adagolodjo2019}; instead, the overall control system is evaluated, with model errors compensated by high-frequency non-rigid tissue registration enabled by six Flex13 cameras, which is an unrealistic requirement in actual clinical settings.

We present 1) a mechanics-based needle-tissue interaction model for bevel-tip needle insertion that accounts for nonlinear soft tissue behaviors and multi-layer scenarios, and 2) a real-time, interactive finite element simulation that allows full three-DOF planar control inputs to be placed anywhere along the needle, enabling simulation of various different clinical scenarios. We showcase the versatility of the model and simulation with soft tissue phantoms consisting of up to four different layers, as well as non-homogeneous chicken breast tissues.
% 1) a mechanics-based, real-time simulation that can simulate various bevel tip effects in 2D while taking into account nonlinear tissue behaviors, 2) versatility of the simulation for different clinical scenarios through control inputs that can not only be placed at the needle base, but anywhere along the needle, 3) integration of bevel effects that can be tuned with a single parameter to fit individual needle-tissue pairs, with the possibility to be modified online to simulate active bevel steering and control, 4) ability to directly use in the model tissue mechanical properties obtained from biomechanical research to simulate strain-dependent behaviors in complex multi-layer soft tissue environments, and 5) model validation against phantoms with 4 different layers as well as non-homogeneous chicken breast tissue.
The rest of the paper is structured as follows: Sec.~\ref{sec:modeling_and_simulation} presents bevel-tip needle insertion modeling and simulation. Sec.~\ref{sec:experiment_methods_and_hardware_components} introduces tissue phantoms, hardware and methods used in the experiments. Sec.~\ref{sec:parameter_tuning_and_model_validation} introduces methods used for parameter tuning and model validation, followed by Sec.~\ref{sec:results_and_discussion}, where experiment results are analyzed. Lastly, Sec.~\ref{sec:conclusion_and_future_work} presents a summary of key contributions, research outcomes, and future work prospects. 

\section{Modeling and Simulation}
\label{sec:modeling_and_simulation}

The objective is to create a prototypical mechanics-based bevel tip needle insertion simulation that accounts for 1) the effect of nonlinear strain-dependent behavior of biological soft tissues under compression, 2) the interaction with multiple soft tissue layers of varied mechanical properties, 3) multiple control inputs along the length of the needle, and 4) full three-degree-of-freedom planar needle motions such as insertion, retraction, as well as bending and in-plane rotation. 

Based on the assumptions that the needle length remains unchanged during insertion and the needle deformation is small, the needle is modeled as an inextensile Euler-Bernoulli beam that deforms only due to bending. Static equilibrium of the beam leads to the governing equation
\begin{equation}
  \label{eq:beam_bending}
  EIu_{xxxx}(x) = F(x),
\end{equation}
with $E$ being the elastic modulus of the needle material, $I$ the area moment of inertia of the needle cross-section, $u(x)$ the deformed shape of the needle, and $F(x)$ the distributed load on the needle shaft. Subscript $x$ is used to denote spatial derivatives.

As shown in~\cite{Humphrey2004}, biological soft tissues often exhibit nonlinear stress-stretch response under large strains without a permanent change in structure. Since soft tissues are composites consisting of elements such as elastin, collagen, and water, it is thus suitable to assume a homogenized, macroscopic nonlinear behavior for the development of this work. In~\cite{Wang2023_1}, the authors show that an incompressible, one-term Ogden hyperelastic tissue model paired with unconfined uniaxial compression can be reliably use for predicting the force $F$ generated by the needle bending and rotation through
\begin{align}
  \label{eq:bending_force}
  F(x) & = \sum_{i}f_i(x) \quad x\in\Omega, \\
  \label{eq:individual_force}
  f_i(x) = k_i(x)u(x)&\left\{1 - \gamma_i\sin^2\left[ \tan^{-1} \left( u_{x}(x) \right) \right] \right\}, \:
\end{align}
\begin{equation*}
x \in \Omega_i \text{ s.t. } \forall i \text{ } \Omega_i \subseteq \Omega \text{ }, 
\end{equation*}
where the subscript $i$ relates quantities to the $i^{\text{th}}$ tissue layer. % The set $\Omega$ contains points along the needle shaft across all tissue layers, with $\Omega_i$ being a subset of $\Omega$ where the needle is surrounded by tissue layer $i$.
The quantity in curly brackets accounts for the friction forces between the needle shaft and surrounding tissues, and $\gamma$ is the friction coefficient. In water-rich biological soft tissues, $\gamma \ll  1$, thus~(\ref{eq:individual_force}) can be approximated as
\begin{equation}
  \label{eq:approx_individual_force}
  f_i(x) \approx k_i(x)u(x)\quad x \in \Omega_i,
\end{equation}
where~(\ref{eq:approx_individual_force}) can be interpreted as Hooke's law whose proportionality measure $k_i$ becomes a nonlinear function dependent on the $i^{\text{th}}$ layer position, as well as the current needle deflection $u(x)$ measured from a reference configuration.

For an incompressible one-term Ogden hyperelastic material undergoing unconfined uniaxial compression, its tangent modulus can be evaluated from
\begin{equation}
  \label{eq:tangent_modulus}
  k(\lambda) = \frac{\partial\sigma_{comp}}{\partial\lambda} =  2\mu \left( \lambda^{\alpha-1} + \frac{1}{2}\lambda^{-\frac{\alpha}{2}-1} \right),
\end{equation}
where $\mu$ is the material shear modulus, $\alpha$ a nonlinearity measure~\cite{Lohr2022}, and $\sigma_{comp}$ the stress component in the direction of tissue compression. The stretch variable $\lambda$ is defined as
\begin{equation}
  \label{eq:stretch_ratio}
  \lambda = \frac{ti - |u(x)|}{ti},
\end{equation}
where $ti$ is the tissue's initial thickness prior to compression. An absolute value of $u(x)$ is used in~(\ref{eq:stretch_ratio}) to enforce tissue deformation in the direction of compression, \textit{i.e.} $\lambda \leq 1$. \Cref{eq:beam_bending,eq:bending_force,eq:individual_force,eq:tangent_modulus,eq:stretch_ratio} are solved using finite element method (FEM) by assuming two-node beam elements with cubic Hermite interpolating functions~\cite{Hughes2012}.

Needle insertion and retraction are simulated with a purely geometric approach based on the assumption that the needle is inextensible. Constraint points are added or removed depending on the depth of insertion. Each needle element inserted into the tissue is assigned its closest constraint point, and tissue stretch $\lambda$ is resolved with respect to its assigned constraint point location.

To simulate the needle's bevel effect during insertion, an offset $b$ is applied to new constraint points created at the needle tip (see Fig.~\ref{fig:model_schematic}). This offset can correlate to the tip bevel geometry, and effectively creates a ``pre-stretched spring'' whose reaction force is tissue-dependent. By resolving the needle bending formulation in the next simulation step, the resultant tip force generated by the offset is computationally incorporated into the model, allowing the needle tip to bend in the corresponding direction. By controlling the sign and magnitude of this offset, different bevel effects can be simulated, as shown in Fig.~\ref{fig:model_schematic}. Active needles whose tip orientation is adjustable can also be simulated by changing the offset parameter online, although such use cases are not considered in the current study~\cite{Ryu2011, Karimi2019, Padasdao2021}.

Known coordinates of the needle points can be used as control points along the needle, and are treated as essential boundary conditions for their corresponding node. For example, the needle base can have a known configuration that is robotically controlled; the needle shaft can be constrained to move through a template; the needle tip configuration can be prescribed if tracked by an electromagnetic tracking system. % Because the finite element routine considers force balance on the entirety of the needle, such information can be integrated into the solution routine without affecting the core structure of the simulation. 

To create an interactive simulation, we distinguish two types of user inputs: ``vertical'' $V$-input and ``horizontal'' $H$-input. $V$-input is a condition that directly affects the needle's shape, such as nodal lateral displacement or slope of a particular point on the needle, and is treated similarly to essential boundary conditions of the finite element solver. $H$-input pertains only to needle insertion and retraction, and this kinematic problem is solved independently from the beam-bending problem. 

\begin{algorithm}[b]
\newcommand{\PW}{{}^{\mathcal{W}}}
\newcommand{\PC}{{}^{\mathcal{C}}}\newcommand{\WtoC}{\mathcal{W}\text{to}\mathcal{C}}
\newcommand{\CtoW}{\mathcal{C}\text{to}\mathcal{W}}
\newcommand{\algAnd}{$\,$ \textbf{and} $\,$}
\newcommand{\algOr}{$\,$ \textbf{or} $\,$}
\caption{Interactive bevel needle simulation} \label{alg:workflow}
\begin{algorithmic}[1]
 \Ensure ENV set \Comment{tissue boundaries and needle properties}
  \State initialization, $\PW N, \PW C = \varnothing$ 
  \While {true}
    \State [$V_i^m$, $H_j^m$] $\Leftarrow$ \Call{get\_inputs}{} \Comment{See Fig.~\ref{fig:model_schematic}}
    \If {$\PW C = \varnothing$ \algAnd $x_{tip} \in \Omega_1$}
      \State $\PW C^1 \Leftarrow [x_{tip}, y_{tip}, \theta_{tip}]^\top$
    \EndIf
    \State [$\mathbf{T}_\mathcal{W}^\mathcal{C}, \mathbf{T}_\mathcal{C}^\mathcal{W}$] $\in SE(2)\Leftarrow$ \Call{get\_tf}{$\PW C_1$}
    \State [$\PC N$, $\PC C$] $\Leftarrow$ \Call{apply\_tf}{$\PW N, \PW C, \mathbf{T}_\mathcal{W}^\mathcal{C}$}
      \While {$itr \leq itr_{max}$} \label{ln:inr_begin}
        \State $\PC N \Leftarrow$ \Call{FEM}{$\PC N$, ENV, $V_i^k$} \label{ln:mechanics}
        \If {$converged$}
          \State break
        \EndIf
        \State $itr \Leftarrow itr + 1$ \label{ln:inr_end}
      \EndWhile
    \State [$\PC N^{\prime}, \PC C^{\prime} ]\Leftarrow$ \Call{insertion}{$\PC N$, $H_i^k$} \label{ln:kinematics}
    \State $[\PW N, \PW C ]\Leftarrow$ \Call{apply\_tf}{$\PC N^{\prime}$, $\PC C^{\prime}$, $\mathbf{T}_\mathcal{C}^\mathcal{W}$}
  \EndWhile
\end{algorithmic}
\end{algorithm}

% The simulation is performed in two frames. A fixed $\mathcal{W}$ reference frame defined by the tissue where the simulation is displayed, and a dynamically formed $\mathcal{C}$ frame created by the first constraint point ${}^\mathcal{W}C_1 = [x_{tip}, y_{tip}, \theta_{tip}]^\top$, which is placed at the first contact point between needle tip and tissue boundary. The variable $\theta_{tip}$ is the needle's orientation at the tip. While the simulation is updated in $\mathcal{W}$, the finite element solution is obtained in the $\mathcal{C}$ frame. This allows the needle to be inserted in any initial position and orientation. 

As shown in Fig.~\ref{fig:model_schematic}, the simulation is displayed in a world-fixed frame $\mathcal{W}$, while needle deformation is obtained in a dynamically formulated constraint frame $\mathcal{C}$. After initializing the needle shape ${}^{\mathcal{W}}N$ and soft tissue layers and properties in $\mathcal{W}$, the simulation reads user inputs to update the needle configuration. When contact between needle tip and tissue boundary is detected, the first constraint point ${}^\mathcal{W}C^1 = [x_{tip}, y_{tip}, \theta_{tip}]^\top$ is placed at the needle tip, and two rigid-body transformations $\mathbf{T}_\mathcal{W}^\mathcal{C}$, $\mathbf{T}_\mathcal{C}^\mathcal{W}$ are defined, which are used to transform needle coordinates from frame $\mathcal{W}$ to $\mathcal{C}$ and vice versa. Within step $m$, after including $V_i^m$, the finite element solution is obtained in the $\mathcal{C}$ frame. The converged solution for the needle, ${}^\mathcal{C}N$, is then updated by $H_j^m$ to simulate needle insertion or retraction. A new constraint point can be added with an offset $b$ such that in the next simulation iteration, the needle tip will experience a pulling force in the direction of the bevel, thus creating a bevel insertion effect. Updated needle shape and constraint points are then transformed back to frame $\mathcal{W}$, where the simulation is visually updated. Dynamic formulation of constraint points allows the needle to be inserted in any initial position and orientation. The simulation is written in MATLAB, and the workflow is shown in Algorithm~\ref{alg:workflow}. %A summary of parameters use in the simulation is provided in Table~\ref{tbl:description}. The first four parameters define the needle geometry and properties, which can be obtained from the manufacturer; the next six define soft tissue layers and their properties, which can be measured or obtained from literature on biomechanical research. The last two parameters can be regarded as ``hyperparameters'' that affect constraint points creation and bevel effect.

% \begin{table}[tb]
% \centering
% \caption{Simulation parameters and description.}
% \begin{tblr}{
%   width = \linewidth,
%   colspec = {Q[231]Q[712]},
%   row{1} = {c},
%   cell{2}{1} = {c},
%   cell{3}{1} = {c},
%   cell{4}{1} = {c},
%   cell{5}{1} = {c},
%   cell{6}{1} = {c},
%   cell{7}{1} = {c},
%   cell{8}{1} = {c},
%   cell{9}{1} = {c},
%   cell{10}{1} = {c},
%   cell{11}{1} = {c},
%   cell{12}{1} = {c},
%   hlines,
%   vlines,
%   hline{1,13} = {-}{0.08em},
% }
% \textbf{Parameters}    & \textbf{Description}                             \\
% $E$                    & Needle Young's modulus                           \\
% $L$                    & Needle length                                    \\
% $OD$                   & Needle outer diameter                            \\
% $ID$                   & Needle inner diameter                            \\
% $T_{int}$              & Tissue layer boundaries                          \\
% $\gamma$                    & Needle-tissue friction coefficient               \\
% $\mu$                    & Tissue shear modulus                             \\
% $\alpha$                    & Ogden nonlinearity constant                      \\
% $ti$                   & Tissue length before compression                 \\
% $b$                    & New constraint vertical offset from needle tip   \\
% $C_{int}$              & Horizontal offset between two constraint points             
% \end{tblr}
% \label{tbl:description}
% \end{table}

\section{Experiment Methods and Hardware Components}
\label{sec:experiment_methods_and_hardware_components}
\begin{figure}[b]
  \centering
  \includegraphics[width=\columnwidth]{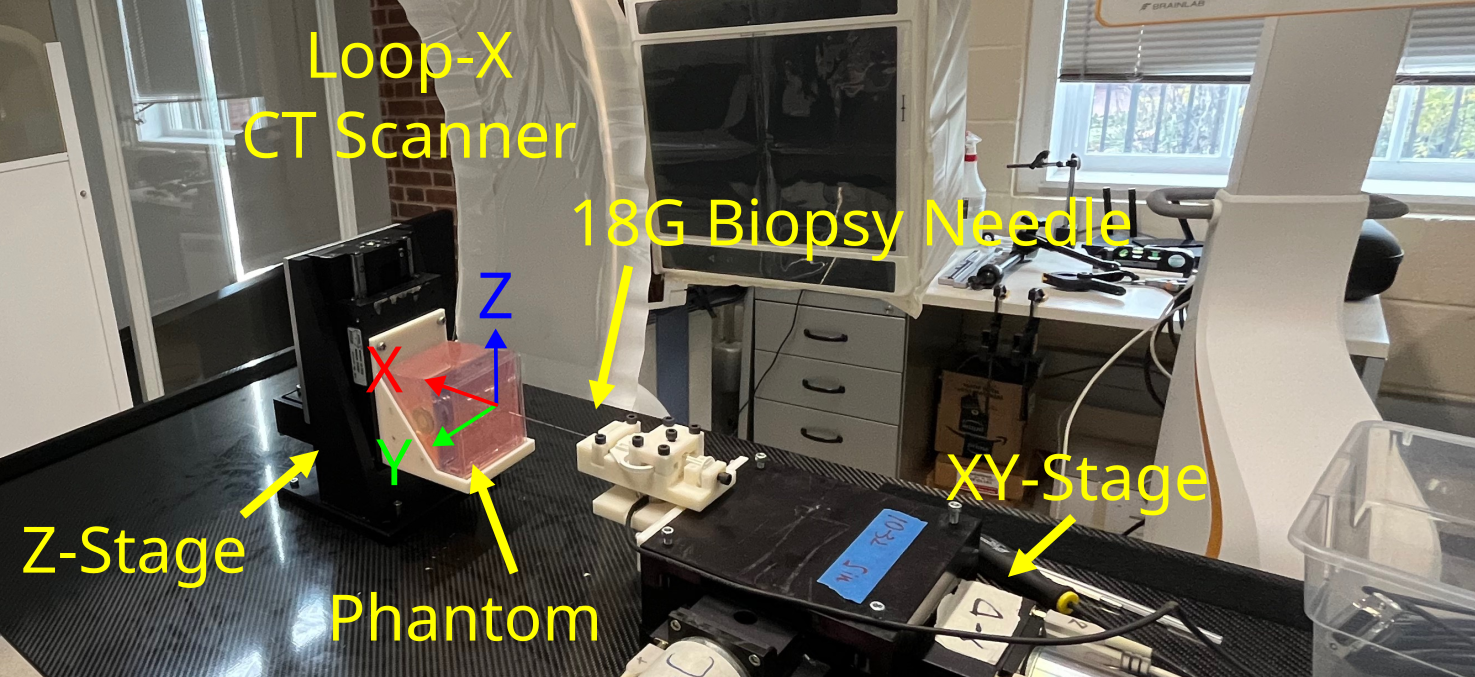}
  \caption{Experiment setup used for validating the simulation.}
  \label{fig:robot}
\end{figure}

To demonstrate the model's ability to match results from physical needle insertion experiments, a set of custom-created multi-layer tissue phantoms with different mechanical properties and geometrical features are developed. These phantoms, in addition to chicken breast tissues, are used for bevel needle insertion experiments, as shown in Fig.~\ref{fig:robot}. Through a cone-beam computed tomography (CT) machine (Loop-X, BrainLab, Germany), scans of the needle after each insertion are obtained for needle shape reconstruction and to tune the model's parameters. Details of each step are presented in the following subsections.

\subsection{Tissue Phantoms}
\label{sec:developed_phantoms}
\begin{figure}[t]
  \centering
  \includegraphics[width=\columnwidth]{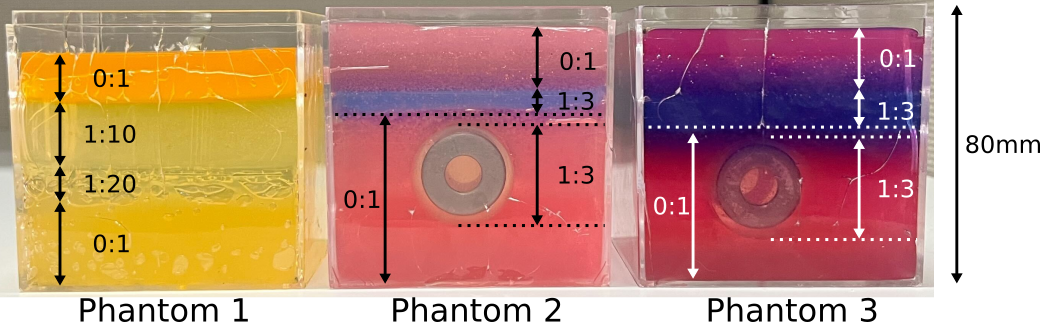}
  \caption{Tissue phantoms used in our experiments, with hardener-to-plastisol volume ratios labeled on each layer.}
  \label{fig:phantoms}
\end{figure}

A total of three different multi-layer phantoms are created, as shown in Fig.~\ref{fig:phantoms}. All phantoms are made of plastisol with varied amounts of hardener added to modify their stiffness. Although other materials are shown to better mimic the mechanical behavior of human soft tissues~\cite{tejo2022soft}, they are still homogeneous in nature, and do not offer substantial advantages to plastisol in the context of this study. Plastisol remains a viscoelastic material, and considering the relatively small amount of strain the layers undergo during the needle insertion process in our experimental setup, a material that can withstand substantial elastic deformation is of secondary importance.
% \textcolor{blue}{Although larger than desired, the friction coefficient between the needle shaft and plastisol is implicitly embedded within the tunable hyper-parameters of the model, and thus bears little effect on the ``realism" of the phantom. -- Can we explain this a little more? Not sure if I completely understant this statement}

We use two types of plastisol: PM242 Ultra Clear/Low Odor Medium Plastisol (Bait Plastics, USA) and Regular Plastic (M-F Manufacturing, USA), along with their corresponding hardeners. In its original form, the liquid plastisol is white in color that turns translucent when heated to around $250^{\circ}$C. The mixture solidifies after cooling, and layering effect is achieved by sequentially casting heated mixtures after a previously solidified layer. All phantoms are cast in $80\times80\times80$~mm$^3$ acrylic boxes to reduce boundary slippage. Resin dye is added to visually distinguish each layer and facilitate layer thickness measurements.

Phantom 1 consists of four layers of varied thicknesses and mechanical properties and is developed for testing an insertion scenario with simple 2D geometry. Phantoms 2 and 3 are designed with the clinical context of prostate biopsies in consideration. Their first layer represents the different fat and muscle tissues encountered by the needle at the perineum entry point. Their second layer represents the diaphragm, which is known to be responsible for substantial needle deflection during biopsy. Their third layer represents the remaining soft tissues separating the diaphragm from the prostate, and their fourth ``layer'' is the prostate itself, represented here as a cylinder. The qualitative relative stiffness of aforementioned layers is chosen to mimic relative stiffness found in human anatomy \cite{boubaker2009finite}. % Hardener-to-plastisol ratios for each layer are reported in Table~\ref{tbl:ratios}.

\subsection{Hardware Setup} 
\label{sec:experiment_setup}
An 18 gauge, 15 cm long, notched biopsy needle (Cook Medical, USA) is employed for all experiments. To ensure a controlled needle insertion, a two-stage setup is used, as shown in Fig.~\ref{fig:robot}. The needle is rigidly affixed to an XY linear stage (XY-6060, Danaher Precision Systems, USA), which is used to advance the needle forward and laterally. % A 6 degrees of freedom force/torque sensor (ATI Nano17, ATI industrial automation) is positioned underneath the needle holder to collect force information during experiments.To take better advantage of the prostate phantoms' geometry,
A Z linear stage (Model 12-1532, Dover Motion, USA) is used for adjusting the phantoms' vertical position to vary the type of tissue layers the needle can traverse and their effective thickness. The stages are secured on an acrylic plate, which also allows the positioning of the Z-stage at different angles. If the phantom's insertion surface normal to the needle is defined at $0^{\circ}$, the Z-stage can be placed at $\pm 10^{\circ}$, $\pm 20^{\circ}$, and $\pm 30^{\circ}$, further expanding possible setup configurations. %No needle template is used in the present case.

\subsection{Image Processing and Needle Reconstruction}
\label{sec:image_processing_and_needle_reconstruction}
% Obtaining the needle's shape accurately after insertion presents significant challenges. T
In order to avoid optical distortions caused by the simulant material, % and the various distortions arising from the diverse layers integrated into the phantoms greatly undermine the effectiveness of vision-based needle identification. Efforts to align cameras and track the 3D needle position are largely susceptible to errors. Therefore, we referred to
CT scans with a voxel size of $0.46\times0.46\times0.46$~mm$^{3}$ are used to extract the needle's shape after each insertion, as well as the effective insertion depth. Intensity thresholding is applied to the acquired images to segment the needle in 3D; radiopaque markers are used to transform the reconstructed needle to the world-fixed frame $\mathcal{W}$ using the Iterative Closest Point algorithm.  
% This invovled segmenting the needle from the CT images using basic thresholding techniques and geometric context. It is important to note that the voxel size stands at $0.46$ mm, which does introduce limitations, but represents the optimal resolution achievable.
% Given the absence of a restraining template, the needle's deflection within a 2D plane is improbable; instead, the needle is likely to experience out-of-plane bending away from the bevel direction. Considering the focus of this paper on planar needle simulation, it is essential to address these aspects to compare simulation outcomes to experimental results. After measuring layer thicknesses $T_i$, we calculate the boundaries of distinct layers along various needle points, represented as red dots in Fig.~\ref{fig:needle_bending}. Subsequently, we approximate the effective tissue thickness $T_i'$ that the needle traversed by measuring the $L_2$ norm between two consecutive red dots in the shown planar view, which is used in our comparisons.
Knowing the relative position of the reconstructed needle with respect to the tissue phantom and the thickness of each layer of the phantom, the effective distance traversed by the needle on the bevel bending plane is calculated and used to reconstruct the tissue boundaries in our planar simulation.

%\begin{figure}[b]
%  \centering
%  \includegraphics[width=\columnwidth]{./figures/3d_to_2d.png}
%  \caption{Extraction of effective layers' thickness to compensate for out-of-plane needle bending when comparing to 2D simulation outcomes. Needle bending is exaggerated for visual purposes only. \textcolor{blue}{probably need an actual 3D figure. the current 3D might still be confusing for those unfamiliar of our work. I will work on it. SolidWorks might do.}}
%  \label{fig:needle_bending}
%\end{figure}

\section{Parameter Tuning and Model Validation}
\label{sec:parameter_tuning_and_model_validation}
Due to the physical nature of the proposed model, parameters need to be tuned to align the simulation to the empirical data. We consider tuning for constraint offset $b$, as well as material shear modulus $\mu$ and Ogden nonlinearity constant $\alpha$ for each layer, although the latter can be obtained via mechanical testing~\cite{Singh2021}.

\subsection{Model Parameters Tuning}
\label{sec:needle_insertion_experiments}
Tuning experiments aim to examine the following aspects: 
\begin{itemize}
\item Model validity across different numbers of layers traversed by the needle.
\item Model validity across different tissue phantoms, as well as heterogeneous animal tissues.
\end{itemize}

For the first aspect, Phantom 2 is selected for conducting multiple insertions at various depths involving variations in insertion position and orientation. This approach effectively yields needle insertions spanning distinct layer counts and different layer thicknesses within the same phantom. For the second aspect, all three phantoms are used for carrying out additional insertions at various depths, as well as experiments on 3 pieces of chicken breast. One of the chicken breasts is punctured at 2 different orientations, resulting in a total of $3\times4 = 12$ insertions. Model parameters are manually tuned for each phantom. The media used, traversed layers, number of insertions, and the depth range are reported in Table~\ref{tbl:layers}.

Both needle tip and needle shape prediction accuracies are taken into consideration during parameter tuning, and the following metrics are used to evaluate the outcome:
% From a clinical standpoint, accuracy of needle tip position is of significant importance to ensure effective access to specified anatomical targets. However, needle shape prediction accuracy remains important, as it signifies the viability and realism of simulated trajectories. Therefore, the following metrics are considered: 
\begin{itemize}
\item \textbf{Tip Error (TE):} needle tip location error compared to the ground truth
\begin{equation} \label{eq7}
TE = ||\boldsymbol{r}^K_{sim} - \boldsymbol{r}^K_{CT}||_2
\end{equation}
\item \textbf{In-Plane Error (IPE):} shape error at each point $j$ of the needle in the natural bending plane of the bevel-tip
\begin{equation}
IPE_j = ||\boldsymbol{r}^j_{sim} - \boldsymbol{r}^j_{CT} ||_2
\end{equation}
\item \textbf{Error-to-Deflection Percentage (EDP):} the percentage ratio of the tip error computed in Eq.~\ref{eq7} to the needle's tip deflection extracted from CT data
\begin{equation}
EDP = \frac{TE}{|{}^{\mathcal{W}}N^K_y - {}^{\mathcal{W}}N^1_y|} \times 100
\end{equation}
\end{itemize}
where $\boldsymbol{r} = [{}^{\mathcal{W}}N^1_x,  {}^{\mathcal{W}}N^1_y,..., {}^{\mathcal{W}}N^K_x, {}^{\mathcal{W}}N^K_y]^\top$ with $K$ being the total number of discretized needle points from the CT ground truth, and $\boldsymbol{r}^j = [{}^{\mathcal{W}}N^j_x,  {}^{\mathcal{W}}N^j_y]^\top$.

% Additionally, to better illustrate the relationship between reported errors and magnitude of needle tip deflection, the following metric is used for analysis:

%1) the absolute error along along the entire needle length, which is the disparity between the simulated needle shape and the actual needle shape as extracted from experiments, and 2) the absolute error specifically at the needle tip.

\begin{table}[t]
\caption{Summary of needle insertion experiments.}
\label{tbl:layers}
\resizebox{\columnwidth}{!}{%
\begin{tabular}{|l|c|c|c|}
\hline
\multicolumn{1}{|c|}{\textbf{Medium}}     & \textbf{Layers} & \textbf{Insertions} & \textbf{Depth Range [mm]} \\ \hline
\multicolumn{1}{|c|}{\textbf{Phantom 1}}  & 3                   & 10                      & 50.69 - 54.27                   \\ \hline
\multirow{4}{*}{\textbf{Phantom 2}}       & 4                   & 9                       & 49.09 - 56.49                   \\ \cline{2-4} 
  & 3                   & 5                       & 33.60 - 54.78                   \\ \cline{2-4} 
  & 2                   & 1                       & 23.79                   \\ \cline{2-4} 
  & 1                   & 1                       & 19.63                  \\ \hline
\multicolumn{1}{|c|}{\textbf{Phantom 3}}  & 4                   & 9                       & 37.68 - 55.32                   \\ \hline
\multicolumn{1}{|c|}{\textbf{Chicken}}    & 1                   & 12                       & 26.98 - 61.22                   \\ \hline
\end{tabular}%
}
\end{table}

\subsection{Model Validation}
\label{sec:model_validation}
To examine the model's generalizability, a series of cross-validation studies are performed (see Table~\ref{table:cross-validation}). These studies aim to resolve the impact of distinct experiment features on the quality of the simulation results.

In Study 1, we use parameters tuned on Phantom 2 with insertions across four layers and examine insertion accuracy within the same phantom but across a reduced number of layers (see Sect.~\ref{sec:parameter_tuning_and_model_validation} for list of parameters). Although the constituents within the same phantom stay the same, the thickness of the traversed layers varies with different insertions. This validation study provides insight into how effectively the model can capture the mechanical properties of different layers, and its ability to transition from complex scenarios to simpler ones. In Study 2, the opposite is performed, \textit{i.e.} we use parameters tuned for two-layer insertions and extend the simulation to a more complex scenario (three and four layers); this study attempts to explain how well the model can transition from a simple scenario (2 layers) to a more complex one (3 and 4 layers) by introducing additional layers with known mechanical properties from the tuned model since the bottom 2 layers in Phantom 2 should have the same mechanical properties as the top 2 ones, as shown in Fig.~\ref{fig:phantoms}.

\begin{table}[b]
\caption{Summary of experiments used for model tuning with corresponding validation studies.}
\label{table:cross-validation}
\resizebox{\columnwidth}{!}{%
\begin{tabular}{c|ccc|ccc|}
\cline{2-7}
  & \multicolumn{3}{c|}{\textbf{Model Tuning}}                                                                       & \multicolumn{3}{c|}{\textbf{Model Validation}}                                                                \\ \hline
\multicolumn{1}{|c|}{\textbf{Study}}              & \multicolumn{1}{c|}{\textbf{Media}}             & \multicolumn{1}{c|}{\textbf{Layers}}    & \textbf{Insertions} & \multicolumn{1}{c|}{\textbf{Media}}             & \multicolumn{1}{c|}{\textbf{Layers}} & \textbf{Insertions} \\ \hline
\multicolumn{1}{|c|}{\multirow{3}{*}{\textbf{1}}} & \multicolumn{1}{c|}{\multirow{3}{*}{Phantom 2}} & \multicolumn{1}{c|}{\multirow{3}{*}{4}} & \multirow{3}{*}{9}   & \multicolumn{1}{c|}{\multirow{3}{*}{Phantom 2}} & \multicolumn{1}{c|}{3}               & 5                    \\ \cline{6-7} 
\multicolumn{1}{|c|}{}                            & \multicolumn{1}{c|}{}                           & \multicolumn{1}{c|}{}                   &                      & \multicolumn{1}{c|}{}                           & \multicolumn{1}{c|}{2}               & 1                    \\ \cline{6-7} 
\multicolumn{1}{|c|}{}                            & \multicolumn{1}{c|}{}                           & \multicolumn{1}{c|}{}                   &                      & \multicolumn{1}{c|}{}                           & \multicolumn{1}{c|}{1}               & 1                    \\ \hline
\multicolumn{1}{|c|}{\multirow{2}{*}{\textbf{2}}} & \multicolumn{1}{c|}{\multirow{2}{*}{Phantom 2}} & \multicolumn{1}{c|}{\multirow{2}{*}{2}} & \multirow{2}{*}{1}   & \multicolumn{1}{c|}{\multirow{2}{*}{Phantom 2}} & \multicolumn{1}{c|}{4}               & 8                    \\ \cline{6-7} 
\multicolumn{1}{|c|}{}                            & \multicolumn{1}{c|}{}                           & \multicolumn{1}{c|}{}                   &                      & \multicolumn{1}{c|}{}                           & \multicolumn{1}{c|}{3}               & 6                    \\ \hline
\multicolumn{1}{|c|}{\textbf{3}}                  & \multicolumn{1}{c|}{Phantom 2}                  & \multicolumn{1}{c|}{3}                  & 1                    & \multicolumn{1}{c|}{Phantom 1}                  & \multicolumn{1}{c|}{3}               & 10                   \\ \hline
\multicolumn{1}{|c|}{\textbf{4}}                  & \multicolumn{1}{c|}{Phantom 2}                  & \multicolumn{1}{c|}{4}                  & 8                    & \multicolumn{1}{c|}{Phantom 3}                  & \multicolumn{1}{c|}{4}               & 9                    \\ \hline
\multicolumn{1}{|c|}{\textbf{5}}                  & \multicolumn{1}{c|}{Chicken}                    & \multicolumn{1}{c|}{1}                  & 3                    & \multicolumn{1}{c|}{Chicken}                    & \multicolumn{1}{c|}{1}               & 9                    \\ \hline
\end{tabular}%
}
\end{table}

Studies 3 and 4 examine the model's capability to generalize to different phantoms. In Study 3, the model is tuned on insertions across three layers in Phantom 2, but validated on insertions across three layers in Phantom 1. The two phantoms are known to have different mechanical properties by design. As for Study 4, we use Phantoms 2 and 3 for tuning and validation, respectively, whereby the two phantoms are designed to be similar in terms of geometrical features as well as mechanical properties, with a total of four traversed layers by the needle.

Study 5 is designed to test the model's ability to adapt to non-homogeneous animal tissues such as chicken breasts. Data collected on the first chicken breast (3 insertions) is selected for parameter tuning, and the tuned model is then tested against the remaining insertion experiments on the two other chicken breasts.

\section{Results and Discussion}
\label{sec:results_and_discussion}

\subsection{Model Parameters Tuning}
\label{sec:multilayer_insertion_validation}

The results of the needle insertion experiments for the parameters' tuning are illustrated in Table~\ref{tbl:results}. The needle's Young's Modulus $E$ is set to be 80 GPa, and the simulation runs steadily with 1mm elements at about 80 Hz on an Intel Core i7-9750H CPU. Tissue parameters are adjusted proportionally to reflect the relative mechanical properties of the phantom's distinct layers. %, which testifies for the model's realism and physical accuracy. 
The numerical values for the model's parameters for all experiments are reported in Table~\ref{tbl:params}. Registration error for the needle reconstruction is found to be on average 0.19 mm. Voxel size combined with said registration error could contribute to observed discrepancies between real and simulated results.

\begin{table}[b]
\caption{Error statistics using the tuned models across different experiments. Units in mm except for EDP.}
\label{tbl:results}
\resizebox{\columnwidth}{!}{%
\begin{tabular}{cc|cccc|}
\cline{3-6}
\multicolumn{1}{l}{} &  & \multicolumn{4}{c|}{\textbf{Average Values}} \\ \hline
\multicolumn{1}{|c|}{\textbf{Layers}} & \textbf{Median Err} & \multicolumn{1}{c|}{\textbf{IPE}} & \multicolumn{1}{c|}{\textbf{Max IPE}} & \multicolumn{1}{c|}{\textbf{TE}} & \textbf{EDP} \\ \hline
\multicolumn{1}{|c|}{\textbf{4}} & 0.18 & \multicolumn{1}{c|}{0.23$\pm$0.17} & \multicolumn{1}{c|}{0.62} & \multicolumn{1}{c|}{0.44} & 10.3 \\ \hline
\multicolumn{1}{|c|}{\textbf{3}} & \textbf{0.16} & \multicolumn{1}{c|}{\textbf{0.18$\pm$0.13}} & \multicolumn{1}{c|}{\textbf{0.49}} & \multicolumn{1}{c|}{\textbf{0.16}} & \textbf{4.77} \\ \hline
\multicolumn{1}{|c|}{\textbf{2}} & 0.18 & \multicolumn{1}{c|}{0.20$\pm$0.14} & \multicolumn{1}{c|}{0.53} & \multicolumn{1}{c|}{0.53} & 58.1 \\ \hline
\multicolumn{1}{|c|}{\textbf{1}} & 0.36 & \multicolumn{1}{c|}{0.36$\pm$0.14} & \multicolumn{1}{c|}{0.60} & \multicolumn{1}{c|}{0.57} & 135 \\ \hline
\textbf{} & \textbf{} & \multicolumn{4}{c|}{\textbf{}} \\ \hline
\multicolumn{1}{|c|}{\textbf{Media}} & \textbf{Median Err} & \multicolumn{1}{c|}{\textbf{IPE}} & \multicolumn{1}{c|}{\textbf{Max IPE}} & \multicolumn{1}{c|}{\textbf{TE}} & \textbf{EDP} \\ \hline
\multicolumn{1}{|c|}{\textbf{Phantom 1}} & 0.21 & \multicolumn{1}{c|}{0.24$\pm$0.18} & \multicolumn{1}{c|}{0.56} & \multicolumn{1}{c|}{0.29} & 26.6 \\ \hline
\multicolumn{1}{|c|}{\textbf{Phantom 2}} & \textbf{0.17} & \multicolumn{1}{c|}{\textbf{0.21$\pm$0.16}} & \multicolumn{1}{c|}{0.57} & \multicolumn{1}{c|}{0.37} & 19.4 \\ \hline
\multicolumn{1}{|c|}{\textbf{Phantom 3}} & 0.22 & \multicolumn{1}{c|}{0.26$\pm$0.19} & \multicolumn{1}{c|}{0.60} & \multicolumn{1}{c|}{\textbf{0.18}} & \textbf{16.5} \\ \hline
\multicolumn{1}{|c|}{\textbf{Chicken}} & 0.22 & \multicolumn{1}{c|}{0.27$\pm$0.18} & \multicolumn{1}{c|}{\textbf{0.54}} & \multicolumn{1}{c|}{0.35} & 76.9 \\ \hline
\end{tabular}%
}
\end{table}

\subsubsection{Testing model against multiple layers}
From Table~\ref{tbl:results}, the median as well as average IPE errors across all four layers in Phantom 2 are within 0.37 mm. The largest average IPE error per insertion is 0.62 mm for the four-layer insertion, and the largest average TE observed for the one-layer insertion is 0.57 mm. The EDP values for deeper insertions are smaller than those for shallower insertions, which is expected since shallower insertions (such as the case for a one-layer insertion) result in smaller needle deflection -- an error as small as 0.3 mm can translate to a relatively large EDP if the needle deflection is below 0.5 mm. Keeping in mind clinical applications for biopsy needle insertions, a submillimeter accuracy would be acceptable nonetheless, in particular for prostate biopsies with average insertion depths of 80 mm \cite{moreira2018evaluation}. The results demonstrate that the simulation can successfully and consistently match physical experiments across various number of layers. % The effect of boundary conditions, however, between the different layers in the phantom, cannot be investigated in this study. This is due to the nature of the phantom creation process, where the cured layer partially melts upon the addition of fresh simulant, resulting in effective cohesion between the layers.

\subsubsection{Testing model against different media} 
\label{sec:multimedium_insertion_validation}
To assess the simulation performance across various phantoms, we tuned the model parameters for each media to align the simulation outcomes with experiment data. These findings are detailed in Table~\ref{tbl:results}. Both median and average errors do not exceed 0.27 mm, with a maximum average IPE of 0.60 mm observed for Phantom 3. The average needle TE is also smaller than 0.36 mm, with small EDP percentages across all tested phantoms. The model is able to match the data collected from the heterogeneous chicken tissues without degradation in performance. The large EDP value for the chicken experiments is a result of the small observed needle deflection from the CT scans due to shallower needle insertions. The results show that the model is capable of simulating different insertions into the same media, irrespective of the number of layers and their corresponding thickness. 

\begin{table}[b]
\caption{Results of the different model validation studies. Units in mm except for EDP.}
\label{tbl:validation}
\resizebox{\columnwidth}{!}{%
\begin{tabular}{ccc|cccc|}
\cline{4-7}
\multicolumn{1}{l}{} & \multicolumn{1}{l}{} &  & \multicolumn{4}{c|}{\textbf{Average Values}} \\ \hline
\multicolumn{1}{|c|}{\textbf{Study}} & \multicolumn{1}{l|}{\textbf{Layers}} & \textbf{Median Err} & \multicolumn{1}{c|}{\textbf{IPE}} & \multicolumn{1}{c|}{\textbf{Max IPE}} & \multicolumn{1}{c|}{\textbf{TE}} & \textbf{EDP} \\ \hline
\multicolumn{1}{|c|}{\multirow{3}{*}{\textbf{1}}} & \multicolumn{1}{c|}{\textbf{3}} & 0.17 & \multicolumn{1}{c|}{0.19$\pm$0.13} & \multicolumn{1}{c|}{0.54} & \multicolumn{1}{c|}{0.14} & 5.56 \\ \cline{2-7} 
\multicolumn{1}{|c|}{} & \multicolumn{1}{c|}{\textbf{2}} & 0.20 & \multicolumn{1}{c|}{0.22$\pm$0.15} & \multicolumn{1}{c|}{0.59} & \multicolumn{1}{c|}{0.59} & 64.8 \\ \cline{2-7} 
\multicolumn{1}{|c|}{} & \multicolumn{1}{c|}{\textbf{1}} & 0.39 & \multicolumn{1}{c|}{0.39$\pm$0.15} & \multicolumn{1}{c|}{0.64} & \multicolumn{1}{c|}{0.64} & 151 \\ \hline
\multicolumn{1}{|c|}{\multirow{2}{*}{\textbf{2}}} & \multicolumn{1}{c|}{\textbf{4}} & 0.24 & \multicolumn{1}{c|}{0.28$\pm$0.22} & \multicolumn{1}{c|}{0.77} & \multicolumn{1}{c|}{0.50} & 11.7 \\ \cline{2-7} 
\multicolumn{1}{|c|}{} & \multicolumn{1}{c|}{\textbf{3}} & 0.19 & \multicolumn{1}{c|}{0.24$\pm$0.19} & \multicolumn{1}{c|}{0.62} & \multicolumn{1}{c|}{0.43} & 12.0 \\ \hline
\multicolumn{1}{|c|}{\textbf{3}} & \multicolumn{1}{c|}{\textbf{3}} & 1.41 & \multicolumn{1}{c|}{1.41$\pm$0.83} & \multicolumn{1}{c|}{2.86} & \multicolumn{1}{c|}{2.82} & 254 \\ \hline
\multicolumn{1}{|c|}{\textbf{4}} & \multicolumn{1}{c|}{\textbf{4}} & 1.09 & \multicolumn{1}{c|}{1.18$\pm$0.68} & \multicolumn{1}{c|}{2.48} & \multicolumn{1}{c|}{2.19} & 201 \\ \hline
\multicolumn{1}{|c|}{\textbf{5}} & \multicolumn{1}{c|}{\textbf{1}} & 0.58 & \multicolumn{1}{c|}{0.46$\pm$0.25} & \multicolumn{1}{c|}{0.85} & \multicolumn{1}{c|}{0.60} & 500 \\ \hline
\end{tabular}%
}
\end{table}

% \begin{table}[b]
% \caption{Needle tip error statistics for insertions through different media. Units in mm.}
% \label{tbl:tip_phantoms}
% \centering
% \begin{tabular}{|c|c|c|c|c|c|}
% \hline
% \textbf{Media} & \textbf{Avg Err} & \textbf{Std} & \textbf{Med Err} & \textbf{Min Err} & \textbf{Max Err} \\ \hline
% \textbf{Phantom 1}      &         0.289         &         0.071     &       0.281           &                0.155  &             0.391     \\ \hline
% \textbf{Phantom 2}      &          0.366        &          0.242    &  0.271                &              0.033    &      0.724            \\ \hline
% \textbf{Phantom 3}      &     0.187             & 0.113            &       0.186           &      0.028            &         0.349         \\ \hline
% \textbf{Chicken}      &        0.385          & 0.211           &            0.292      &   0.184        &           0.759       \\ \hline
% \end{tabular}
% \end{table}

\subsection{Model Validation}
\begin{figure}[t]
  \centering
  \includegraphics[width=\columnwidth]{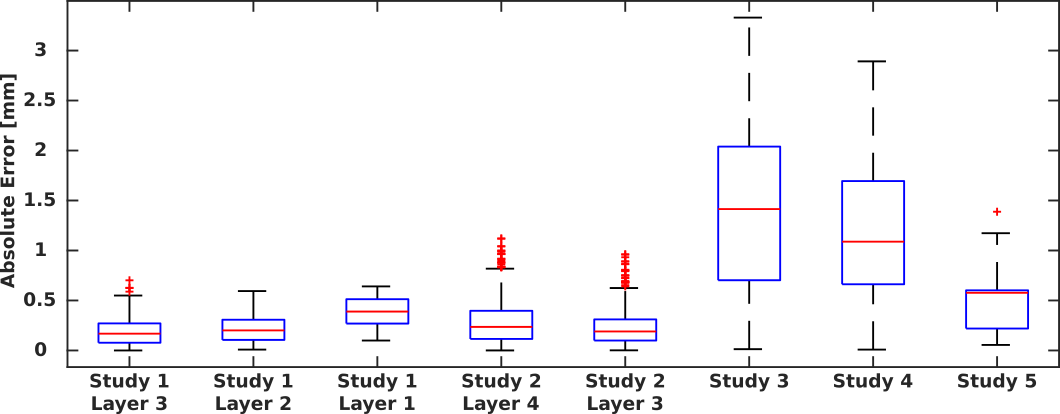}
  \caption{Boxplots for all conducted validation studies.}
  \label{fig:valid}
\end{figure}

The results for the model validation studies are reported in Table~\ref{tbl:validation} and Fig.~\ref{fig:valid}, with corresponding model parameters in Table~\ref{tbl:params}. In both Studies 1 and 2, the largest median and average IPE errors are below 0.39 mm, with the largest average maximum IPE per insertion being 0.77 mm, and largest average tip error 0.59 mm. This shows that the model can be tuned on both complex ($>2$ layers) and simpler (2 layers in this case) models, without loss of generalizability when transitioning to different insertions within the same medium. It is important to highlight the clinical benefit of such observations, whereby only a handful of insertions into a patient could be used to reasonably predict the needle shape and deflection for different insertions with the same patient.

In Studies 3 and 4, the overall errors are larger compared to Studies 1 and 2, with the largest average maximum IPE reaching 2.86 mm in Study 3. The results do, however, align with our expectations. Phantom 1 is inherently different from Phantom 2 in terms of mechanical properties, so parameter values identified for a model that fits Phantom 2 are unlikely to immediately translate to a different phantom. Similarly, although Phantom 3 is designed to replicate Phantom 2, it is created by two different people at different times, which implies that some differences between the two will still be observed. In a real scenario, these differences represent patient-to-patient variability, and underscore the futility of seeking a ``one-fits-all'' solution that will seamlessly work for all patients without modification.

The submillimeter errors reported in Study 5 highlight the model's potential to reasonably capture complex heterogeneous tissue-needle interactions. Fig.~\ref{fig:valid} demonstrates the model's capability to generalize within the same medium and across heterogeneous tissues, but not so much across different types of phantoms.

The large EDP values in Table~\ref{tbl:validation} (>100\%) across all studies are attributed to the smaller needle deflections (0.02-1.5 mm) due to shallower insertions, as well as the smaller amount of data available for analysis in certain cases, such as in Study 1 for the two and one-layer insertions.

 It is important to note the limitations of the study. First, the parameters are manually tuned, which means that a more robust optimization strategy will likely lead to even better results. Second, some studies only had one needle insertion experiment (Phantom 2 with two-layer and one-layer insertions), which could skew some of the reported results. Lastly, the realism of the phantoms compared to human tissues is limited.

\begin{table}[b]
\caption{Model parameters tuned for each set of experiments. Ph: Phantom, Ch: Chicken, T: Table, St: Study.}
\label{tbl:params}
\resizebox{\columnwidth}{!}{%
\begin{tabular}{l|cc|cc|cc|cc|c|}
\cline{2-10}
\multicolumn{1}{c|}{}                     & \multicolumn{2}{c|}{\textbf{Layer 1}}                   & \multicolumn{2}{c|}{\textbf{Layer 2}}                   & \multicolumn{2}{c|}{\textbf{Layer 3}}                   & \multicolumn{2}{c|}{\textbf{Layer 4}}                   & \textbf{Bevel} \\ \cline{2-10} 
\multicolumn{1}{c|}{}                     & \multicolumn{1}{c|}{\textbf{$\mu$}} & \textbf{$\alpha$} & \multicolumn{1}{c|}{\textbf{$\mu$}} & \textbf{$\alpha$} & \multicolumn{1}{c|}{\textbf{$\mu$}} & \textbf{$\alpha$} & \multicolumn{1}{c|}{\textbf{$\mu$}} & \textbf{$\alpha$} & \textbf{$b$}   \\ \hline
\multicolumn{1}{|l|}{\textbf{Ph2 T4}}     & \multicolumn{1}{c|}{2e5}            & 1                 & \multicolumn{1}{c|}{3.3e7}          & -1                & \multicolumn{1}{c|}{2e5}            & 1                 & \multicolumn{1}{c|}{3.3e7}          & -1                & 0.085          \\ \hline
\multicolumn{1}{|l|}{\textbf{Ph1 T5}}     & \multicolumn{1}{c|}{2e5}            & 1                 & \multicolumn{1}{c|}{3.3e7}          & -1                & \multicolumn{1}{c|}{2e6}            & 1                 & \multicolumn{1}{c|}{3.3e7}          & -1                & 0.03           \\ \hline
\multicolumn{1}{|l|}{\textbf{Ph2 T5}}     & \multicolumn{1}{c|}{2e5}            & 1                 & \multicolumn{1}{c|}{3.3e7}          & -1                & \multicolumn{1}{c|}{2e5}            & 1                 & \multicolumn{1}{c|}{3.3e7}          & -1                & 0.085          \\ \hline
\multicolumn{1}{|l|}{\textbf{Ph3 T5}}     & \multicolumn{1}{c|}{2e5}            & 1                 & \multicolumn{1}{c|}{3.3e7}          & -1                & \multicolumn{1}{c|}{2e5}            & 1                 & \multicolumn{1}{c|}{3.3e7}          & -1                & 0.03           \\ \hline
\multicolumn{1}{|l|}{\textbf{Ch T5}}      & \multicolumn{1}{c|}{1e3}            & 1                 & \multicolumn{1}{c|}{n/a}            & n/a               & \multicolumn{1}{c|}{n/a}            & n/a               & \multicolumn{1}{c|}{n/a}            & n/a               & 0.085          \\ \hline
\multicolumn{1}{|l|}{\textbf{Ph2 T6 St1}} & \multicolumn{1}{c|}{2.2e5}          & 1                 & \multicolumn{1}{c|}{3.2e7}          & -1                & \multicolumn{1}{c|}{2.2e5}          & 1                 & \multicolumn{1}{c|}{3.2e7}          & -1                & 0.085          \\ \hline
\multicolumn{1}{|l|}{\textbf{Ph2 T6 St2}} & \multicolumn{1}{c|}{5e4}            & 1                 & \multicolumn{1}{c|}{1e7}            & -1                & \multicolumn{1}{c|}{5e4}            & 1                 & \multicolumn{1}{c|}{1e7}            & -1                & 0.085          \\ \hline
\multicolumn{1}{|l|}{\textbf{Ph2 T6 St3}} & \multicolumn{1}{c|}{2.2e5}          & 0.85              & \multicolumn{1}{c|}{3.2e7}          & -0.98             & \multicolumn{1}{c|}{2.2e5}          & 0.85              & \multicolumn{1}{c|}{3.2e7}          & -0.98             & 0.085          \\ \hline
\multicolumn{1}{|l|}{\textbf{Ph2 T6 St4}} & \multicolumn{1}{c|}{2.2e5}          & 1                 & \multicolumn{1}{c|}{3.2e7}          & -1                & \multicolumn{1}{c|}{2.2e5}          & 1                 & \multicolumn{1}{c|}{3.2e7}          & -1                & 0.085          \\ \hline
\multicolumn{1}{|l|}{\textbf{Ch T6 St5}}  & \multicolumn{1}{c|}{4e3}            & 0.2               & \multicolumn{1}{c|}{n/a}            & n/a               & \multicolumn{1}{c|}{n/a}            & n/a               & \multicolumn{1}{c|}{n/a}            & n/a               & 0.085          \\ \hline
\end{tabular}%
}
\end{table}

\section{Conclusion and Future Work}
\label{sec:conclusion_and_future_work}

In this paper, a mechanics-based bevel-tip needle model is successfully developed, simulated, and validated against experiments in phantoms and chicken breast tissues. The model has demonstrated its capability to replicate the needle behaviors in actual physical experiments. Additionally, the proposed model has proven to be effective with submillimeter prediction accuracy both on needle tip position and overall shape in scenarios where generalization is justifiable, such as extending its utility across consistent media or relatively similar heterogeneous tissues.

Experiment results show great potential for future work development, namely extending the model to 3D applications, developing simulation-based controllers for trajectory tracking, as well as integrating needle shape feedback modalities such as intraoperative imaging, electromagnetic tracking, and fiber optic sensing.

% \section*{Appendix}

% Appendixes go here
% \section*{Acknowledgment}
% Acknowledgment goes here

\bibliography{bibliography}
\bibliographystyle{ieeetr}

\end{document}